\documentclass{article}
\usepackage{placeins}

\usepackage{PRIMEarxiv}

\usepackage[utf8]{inputenc} 
\usepackage[T1]{fontenc}    
\usepackage{hyperref}       
\usepackage{url}            
\usepackage{booktabs}       
\usepackage{amsfonts}       
\usepackage{nicefrac}       
\usepackage{microtype}      
\usepackage{lipsum}
\usepackage{fancyhdr}       
\usepackage{graphicx}       
\graphicspath{{media/}}    

\usepackage{amsmath}
\usepackage{tcolorbox}  

\usepackage{colortbl}

\title{LLMs in Disease Diagnosis: A Comparative Study of DeepSeek-R1 and O3 Mini Across Chronic Health Conditions}

\author{
  Gaurav Kumar Gupta \\
  Youngstown State University \\
  Youngstown, OH \\
  \texttt{gkgupta@student.ysu.edu} \\
  \And
  Pranal Pande \\
  Youngstown State University\\
  Youngstown, OH \\
  \texttt{ppande@student.ysu.edu} \\
  \And
  Nirajan Acharya \\
  Youngstown State University \\
  Youngstown, OH \\
  \texttt{nirajanach3@gmail.com} \\
  \And
  Aniket Kumar Singh \\
  Youngstown State University \\
  Youngstown, OH \\
  \texttt{aniketkashyyap@gmail.com} \\
  \And
  Suman Niroula \\
  Youngstown State University \\
  Youngstown, OH \\
  \texttt{sum.nir1@gmail.com} \\
}

\begin{document}
\maketitle

\begin{abstract}
Large Language Models (LLMs) are revolutionizing medical diagnostics by enhancing both disease classification and clinical decision-making. In this study, we evaluate the performance of two LLM-based diagnostic tools, \textbf{DeepSeek R1} and \textbf{O3 Mini}, using a structured dataset of symptoms and diagnoses. We assessed their predictive accuracy at both the disease and category levels, as well as the reliability of their confidence scores. DeepSeek R1 achieved a disease-level accuracy of 76\% and an overall accuracy of 82\%, outperforming O3 Mini, which attained 72\% and 75\% respectively. Notably, DeepSeek R1 demonstrated exceptional performance in Mental Health, Neurological Disorders, and Oncology, where it reached 100\% accuracy, while O3 Mini excelled in Autoimmune Disease classification with 100\% accuracy. Both models, however, struggled with Respiratory Disease classification, recording accuracies of only 40\% for DeepSeek R1 and 20\% for O3 Mini. Additionally, the analysis of confidence scores revealed that DeepSeek R1 provided high-confidence predictions in 92\% of cases, compared to 68\% for O3 Mini. Ethical considerations regarding bias, model interpretability, and data privacy are also discussed to ensure the responsible integration of LLMs into clinical practice. Overall, our findings offer valuable insights into the strengths and limitations of LLM-based diagnostic systems and provide a roadmap for future enhancements in AI-driven healthcare.
\end{abstract}


\section{Introduction}
Large Language Models (LLMs) have emerged as groundbreaking advancements in artificial intelligence (AI), reshaping various sectors—including healthcare—through their ability to process and generate human-like text. In medical diagnostics, LLMs are being increasingly leveraged to support clinicians by interpreting complex patient data, identifying subtle patterns, and providing diagnostic insights that may otherwise be overlooked. By processing extensive clinical data—from electronic health records and medical literature to patient-reported symptoms—these models offer the potential not only to classify diseases into broad categories but also to predict specific disease names, thereby refining the diagnostic process \cite{zhou2024,wang2024,jamanetwork2023}.

The integration of LLMs into healthcare has the potential to enhance diagnostic processes by improving accuracy, efficiency, and overall clinical decision-making. For instance, LLMs can rapidly analyze unstructured clinical narratives to extract critical diagnostic features and uncover underlying patterns within complex patient data \cite{frontiers2024,ai6010013}. By synthesizing information from diverse sources, these models offer valuable insights that support clinicians in their diagnostic reasoning, ultimately contributing to more informed and effective treatment decisions.

Recent advancements in AI have led to the development of specialized LLM-based platforms tailored for medical applications. Among these, \textbf{DeepSeek R1} and \textbf{O3 Mini} have attracted considerable attention for their robust performance in both automated disease classification and disease name prediction. DeepSeek R1 is designed to capture the nuances of clinical language, enabling it to differentiate between diseases with overlapping symptom profiles with high accuracy \cite{deepseekr1}. Conversely, O3 Mini emphasizes computational efficiency and scalability, making it particularly suitable for rapid diagnostic support in high-volume or resource-constrained clinical settings \cite{o3mini}. Notably, while preliminary findings suggest that DeepSeek R1 may offer superior overall diagnostic accuracy and confidence levels, O3 Mini has shown distinct strengths in certain domains, such as Autoimmune Disease prediction.

Despite these promising developments, the clinical deployment of LLM-based diagnostic tools requires rigorous validation across diverse patient populations and clinical scenarios. It is also crucial to address the ethical implications of these technologies, including potential biases in training data, the opacity of decision-making processes, and stringent data privacy requirements \cite{he2023, nature2024}. Ensuring compliance with data protection standards such as HIPAA and GDPR is essential for maintaining patient trust and safeguarding sensitive information.

In this study, we evaluate the diagnostic performance of DeepSeek R1 and O3 Mini by comparing their ability to classify diseases and predict specific disease names across multiple clinical categories. Our research aims to provide a comprehensive assessment of these LLM-based systems, highlighting their strengths and identifying key areas for improvement. Through this work, we contribute to the ongoing discourse on AI-driven diagnostics and offer insights that could pave the way for more robust, interpretable, and ethically responsible diagnostic tools in healthcare.

\section{Related Work}

Artificial Intelligence (AI) and Large Language Models (LLMs) have been increasingly explored for their role in disease diagnosis, leveraging deep learning techniques to analyze clinical data, electronic health records (EHRs), and symptom descriptions. Several studies have demonstrated the effectiveness of LLMs in disease classification, differential diagnosis, and clinical decision support\cite{frontiers2024}\cite{nature2024}\cite{jamanetwork2023}\cite{dhakal2024gpt4sassessmentperformanceusmlebased}.

A recent scoping review by Zhou et al. (2024) provides a comprehensive analysis of LLM-based methods for disease diagnosis, examining various disease types, clinical datasets, and evaluation techniques \cite{zhou2024}. The study highlights the efficacy of LLMs in diagnostic tasks and underscores the need for standardized benchmarks to ensure fair performance evaluation. Unlike our study, which focuses on model-specific evaluations, their review broadly assesses LLM methodologies and architectures.

Similarly, Wang et al. (2024) investigated the probabilistic medical predictions of LLMs, demonstrating how prompt engineering can enhance the accuracy and flexibility of clinical diagnoses \cite{wang2024}. Their findings suggest that LLMs can dynamically adjust diagnostic predictions based on evolving patient information. In contrast, our study evaluates the confidence scores of specialized AI models, DeepSeek R1 and O3 Mini, across multiple disease categories, providing a structured approach to assessing AI reliability in medical applications. Furthermore, Sun et al. (2024) introduced a conversational AI model that mimics doctor-patient interactions using reinforcement learning to refine diagnostic reasoning \cite{sun2024}. Their system optimizes follow-up questioning and achieves high performance in disease screening and differential diagnosis. Unlike this approach, our study does not involve interactive AI responses but rather evaluates the direct diagnostic accuracy and confidence levels of predefined AI models.

A broader survey by He et al. (2023) examined the deployment of LLMs across various healthcare domains, highlighting key challenges such as fairness, interpretability, accountability, and ethical considerations \cite{he2023}. While their work provides a conceptual overview of LLM applications in healthcare, our study contributes by offering an empirical evaluation of AI model performance in disease classification.

Our prior study, \textit{Digital Diagnostics: The Potential of Large Language Models in Recognizing Symptoms of Common Illnesses} \cite{ai6010013}, focused on analyzing the capabilities of LLMs in diagnosing common illnesses.
 In this work, we expand upon that foundation by classifying chronic diseases, including Diabetes, Cancer, Heart Disease, Mental Health disorders, and Autoimmune Diseases. This shift towards chronic disease classification enables a more structured evaluation of AI model performance in handling complex, long-term medical conditions. Unlike general LLM-based studies that assess language models on broad medical knowledge, our research specifically evaluates DeepSeek R1 and O3 Mini for their ability to classify chronic diseases with high accuracy and confidence. We systematically assess their performance across multiple disease categories, highlighting both their strengths and areas where improvements are needed.

Additionally, this study introduces a confidence-based evaluation approach, ensuring that AI-driven disease classification models not only achieve high accuracy but also provide reliable predictions that can be confidently used in real-world healthcare applications. A key challenge identified in our research is the classification of Respiratory Diseases, an area where both models show relatively lower performance. Addressing such limitations is crucial for enhancing AI adoption in clinical environments.

By conducting a detailed comparison of these AI models and assessing their diagnostic reliability, this study contributes to the ongoing research in AI-driven medical diagnostics, offering practical insights into the optimization and deployment of AI systems in clinical settings. Our findings provide a structured understanding of how AI models perform in chronic disease classification and suggest potential future improvements for their integration into medical practice.

\section{Methodology}

The dataset used in this study was obtained from authoritative medical sources, including \textit{Mayo Clinic}, \textit{WebMD}, \textit{WHO}, \textit{HealthLink BC}, and \textit{Penn Medicine} \cite{healthlinkbc}\cite{mayoclinic}\cite{webmd}. These sources provide verified and widely accepted medical knowledge that serves as a foundation for clinical decision-making and disease diagnosis. The dataset was structured to include comprehensive information on disease categories, specific diseases, and their corresponding symptoms. Each disease was systematically mapped to a set of commonly reported symptoms based on information curated from these medical institutions, ensuring that the dataset represented real-world symptom presentations. Our study specifically targets the classification of chronic diseases, including \textbf{Cancer, Diabetes, Heart Disease, and Mental Health Disorders}. Chronic diseases often present with overlapping symptoms, making accurate classification a challenging yet essential task for improving medical diagnostics.

\subsection{LLMs Models Used}
To conduct this study, we employed two state-of-the-art large language models (LLMs), \textbf{DeepSeek R1} and \textbf{O3 Mini}, which have been developed to enhance disease classification based on symptom analysis. These models were selected due to their advanced capabilities in processing and interpreting medical language, leveraging extensive training data sourced from clinical texts, research publications, and structured medical databases\cite{deepseekr1}\cite{o3mini}.

\textbf{DeepSeek R1} is a sophisticated LLM designed to process medical text efficiently and derive meaningful diagnostic insights. It has been trained on diverse medical literature and symptom-disease mappings, enabling it to predict potential diseases with high accuracy. Its architecture allows for a deep understanding of complex symptom patterns, improving its capability to differentiate between diseases with similar symptom presentations, which is crucial for chronic disease diagnosis \cite{deepseekr1}.

\textbf{O3 Mini}, on the other hand, is a lightweight LLM optimized for real-time medical decision support. While it prioritizes computational efficiency, it still maintains strong diagnostic accuracy, making it suitable for scenarios requiring quick disease classification with minimal resource consumption. O3 Mini integrates knowledge from structured medical databases and symptom assessment algorithms, allowing it to provide concise and interpretable diagnostic predictions. Its efficiency makes it particularly valuable for early detection and continuous monitoring of chronic diseases \cite{o3mini}.

Both models were evaluated by inputting carefully curated symptom sets derived from our dataset. The LLMs processed these inputs and generated outputs that included disease category classification, specific disease predictions, confidence scores, suggested diagnostic steps, and a brief reasoning behind their decision. These outputs were systematically recorded for further analysis, enabling a comprehensive assessment of each model’s performance in chronic disease classification and diagnostic reasoning. By focusing on chronic diseases, this study aims to provide a structured evaluation of how LLMs can enhance medical diagnostics and improve disease prediction accuracy in real-world healthcare applications.

\begin{figure}[h]
    \centering
    \includegraphics[width=1.0\textwidth]{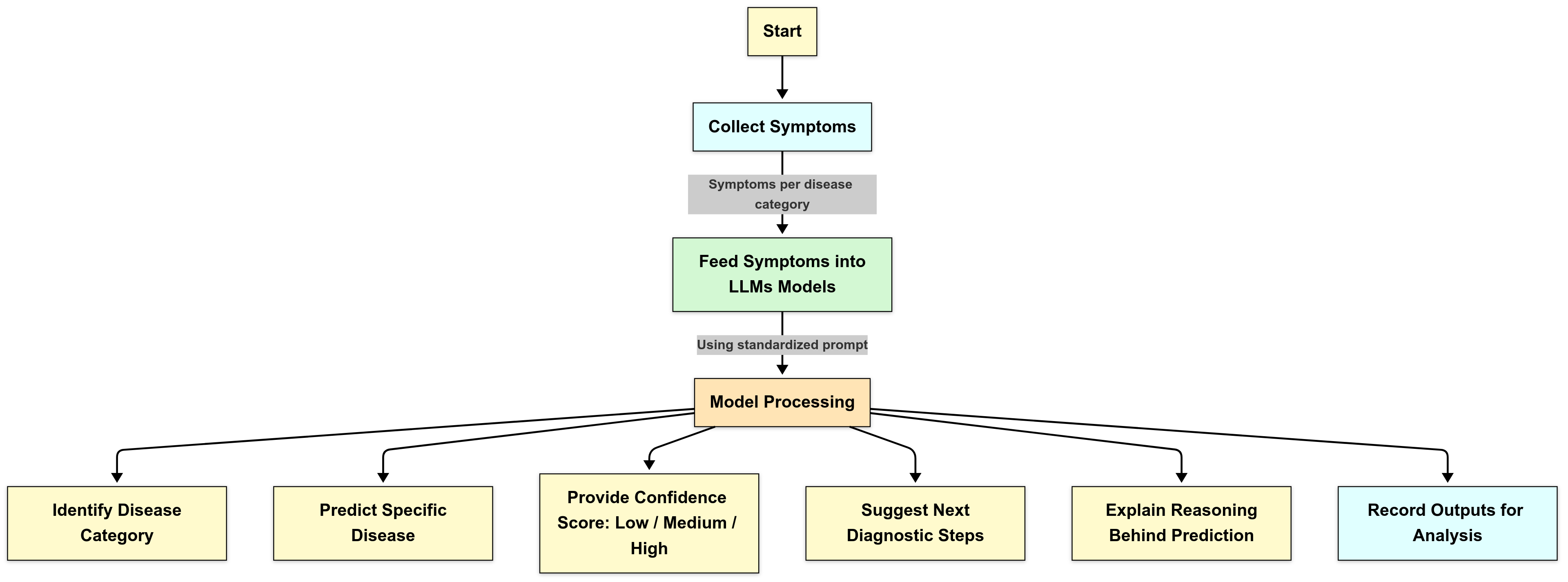}
    \caption{Data Collection Process}
    \label{fig:data}
\end{figure}

\subsection{Data Collection and Testing Process}
To evaluate the performance of the models, we compiled a dataset consisting of 50 distinct symptoms for each disease category, sourced from reputable medical organizations\cite{healthlinkbc}\cite{mayoclinic}\cite{webmd}. These symptoms were carefully curated to ensure a broad and representative coverage of each disease class, reducing potential bias and improving generalizability. The selection process focused on ensuring that both common and less frequent symptoms were included, capturing a realistic distribution of symptom presentations. Given our focus on chronic diseases such as Cancer, Diabetes, Heart Disease, and Mental Health Disorders, the dataset was tailored to reflect the most clinically relevant symptomatology for these conditions.

Once the dataset was finalized, the symptoms were formatted into structured input queries and fed into the LLM models using a standardized prompt. The objective was to simulate real-world diagnostic scenarios where models would be required to analyze symptom patterns and provide meaningful predictions. Each model was tasked with identifying the correct disease category, predicting the most probable disease name, assigning a confidence score (Low/Medium/High), suggesting appropriate next steps for diagnosis or management, and providing a concise explanation of its reasoning. This approach allowed us to capture the interpretability and reliability of the models' predictions, ensuring a robust evaluation framework.

Figure \ref{fig:data} illustrates the structured process of data collection and testing. It begins with symptom collection, followed by data structuring, input processing through LLM models, and output evaluation. Each model processes the standardized symptom inputs and generates diagnostic insights, which are systematically recorded for analysis. The figure outlines key stages, including disease identification, confidence score assignment, and reasoning evaluation, highlighting the step-by-step methodology employed in our study.

The outputs generated by DeepSeek R1 and O3 Mini were systematically recorded for further analysis. These results were compared against the ground truth data to measure classification accuracy, confidence alignment, and diagnostic validity. Additionally, we examined cases where the models exhibited uncertainty or misclassification, providing insights into their limitations and areas for potential improvement. This structured testing and evaluation process enabled a comprehensive assessment of how effectively each LLM could be leveraged for real-world disease classification and clinical decision support. 
Furthermore, we conducted a qualitative analysis of the model outputs to better understand their decision-making pathways. By reviewing model-generated explanations and confidence justifications, we identified patterns in how the LLMs arrived at specific diagnoses. This helped reveal whether certain symptom combinations consistently led to overconfidence or hesitation in prediction. Such insights are critical not only for improving model transparency and trustworthiness but also for highlighting the importance of interpretability when deploying these models in real clinical environments.

\FloatBarrier
\enlargethispage{-3\baselineskip}
\vspace{-8ex}
\subsection{LLMs Evaluation Prompt}
\vspace{-5ex}
The LLM models were assessed using a carefully constructed standardized prompt designed to emulate real-world clinical scenarios. This prompt provided a consistent framework for presenting symptoms, allowing for an objective comparison of the diagnostic performance across different models. The goal was to evaluate each model’s ability to interpret symptom patterns and accurately classify diseases based on the input. The exact prompt used in this study is presented below:
\begin{tcolorbox}[colback=blue!5!white,colframe=red!75!black,title=Prompt for Models]
\textbf{You are an AI medical assistant specializing in disease classification. A patient presents with the following symptoms:}

\textbf{- Symptoms: [Enter symptoms here]}

\textbf{\large Task:}
\begin{enumerate}
\item Classify the disease based on the provided symptoms.
\item Identify the specific disease (predict only one).
\item Provide a confidence score (Low/Medium/High).
\item Suggest next steps (Limit: Maximum 2-3 lines).
\item Explain the reasoning behind the diagnosis (Limit: Maximum 2-3 lines).
\end{enumerate}

\textbf{\large Additional Instructions:}
\begin{itemize}
\item If you are unable to classify the disease with confidence, request a hint.
\item If a hint is needed, ask: \textit{"I need more information. Would you like me to list possible disease categories?"}
\end{itemize}

\textbf{\large Response Format:}
\begin{itemize}
\item \textbf{Disease Classification:} [Predicted category]
\item \textbf{Specific Diagnosis:} [Exact disease name (Only one)]
\item \textbf{Confidence Score:} [Low/Medium/High]
\item \textbf{Next Steps:} [Short recommendation, Max 2-3 lines]
\item \textbf{Reasoning:} [Brief explanation, Max 2-3 lines]
\end{itemize}
\end{tcolorbox}

\subsection{Evaluation and Validation}

In this study, we performed a rigorous quantitative evaluation of the LLMs' diagnostic performance by assessing their ability to accurately predict both the disease category and the specific disease based on a standardized set of symptoms. The evaluation framework is based on a point-based scoring system and a series of quantitative metrics that provide a comprehensive assessment of the model performance.

\subsubsection{Scoring Criteria}
For each test case, the following criteria were used:
\begin{itemize}
    \item \textbf{Disease Prediction:} The LLM receives 1 point if the predicted disease exactly matches the ground truth; otherwise, it receives 0 points.
    \item \textbf{Category Prediction:} The LLM receives 1 point if the predicted disease category corresponds to the correct category; otherwise, it receives 0 points.
\end{itemize}
In addition, the confidence scores provided by the LLMs (categorized as High, Medium, or Low) were recorded for each test case to evaluate the reliability of the predictions.

\subsubsection{Quantitative Metrics}
To quantitatively assess the performance, we defined the following metrics:

\paragraph{1. Disease-Level Accuracy}  
This metric quantifies the percentage of cases in which the LLM correctly predicted the specific disease:
\begin{equation}
\text{Disease-Level Accuracy} = \left( \frac{\sum_{i=1}^{N} \text{Point for Disease}_i}{N} \right) \times 100,
\end{equation}
where \(N\) is the total number of test cases and \(\text{Point for Disease}_i\) is 1 if the prediction for case \(i\) is correct, and 0 otherwise.

\paragraph{2. Category-Level Accuracy}  
This metric measures the proportion of cases in which the LLM correctly classified the disease into its appropriate general category:
\begin{equation}
\text{Category-Level Accuracy} = \left( \frac{\sum_{i=1}^{N} \text{Point for Category}_i}{N} \right) \times 100,
\end{equation}
where \(\text{Point for Category}_i\) is 1 if the category prediction for case \(i\) is correct, and 0 otherwise.

\paragraph{3. Overall Accuracy}  
To provide a holistic view of the LLMs’ performance, we combine the disease-level and category-level accuracies:
\begin{equation}
\text{Overall Accuracy} = \left( \frac{\sum_{i=1}^{N} (\text{Point for Disease}_i + \text{Point for Category}_i)}{2N} \right) \times 100.
\end{equation}
This metric equally weights both the specific disease and category predictions by dividing the total score by \(2N\).

\paragraph{4. Confidence Score Distribution}  
This metric evaluates the reliability of the predictions by examining how frequently each confidence level is assigned. It is computed as:
\begin{equation}
\text{Confidence Score Distribution} = \frac{\text{Count of Cases at a Specific Confidence Level}}{N} \times 100,
\end{equation}
where the numerator represents the number of cases that received a particular confidence level (e.g., High, Medium, or Low) and \(N\) is the total number of test cases.

\subsubsection{Implementation of the Evaluation Framework}
The evaluation was conducted by inputting standardized symptom sets into the LLMs. For each test case, the output—consisting of the disease classification, specific diagnosis, assigned confidence score, and any recommended next steps—was recorded. The point-based scoring system was applied to each test case, and the metrics defined above were computed over the entire test set. This systematic approach allowed for an objective comparison of the LLMs' performance across all symptom sets and disease categories.

 This quantitative evaluation framework ensures a comprehensive and objective assessment of the LLMs by focusing on disease-level accuracy, category-level accuracy, overall accuracy, and the distribution of confidence scores. These metrics provide valuable insights into the strengths and limitations of the LLMs, thereby supporting efforts to enhance diagnostic reliability in clinical applications.

\section{Results}

This section presents the comparative performance analysis of the large language models (LLMs) \textbf{DeepSeek R1} and \textbf{O3 Mini} in disease classification. The analysis is structured based on multiple evaluation criteria, including accuracy metrics, confidence score distribution, and category-specific performance. Figures~\ref{fig:model_performance} and \ref{fig:confidence_analysis} illustrate the key findings, visually depicting accuracy variations across disease categories and confidence levels in model predictions.

\subsection{Overall Performance Metrics}

The evaluation of the models’ diagnostic capabilities was conducted at two levels: disease-level accuracy and category-level accuracy. Disease-level accuracy measures how often a model correctly identifies a specific disease, while category-level accuracy evaluates the ability to classify diseases into broader medical categories.

As summarized in Table~\ref{tab:model_comparison}, DeepSeek R1 demonstrated an overall disease-level accuracy of 76\%, slightly outperforming O3 Mini, which achieved 72\%. Category-level accuracy followed a similar trend, with DeepSeek R1 achieving 88\% and O3 Mini attaining 78\%. When combining disease and category classification accuracy, the overall accuracy of DeepSeek R1 was 82\%, whereas O3 Mini reached 75\%. 

\begin{figure}[h]
    \centering
    \includegraphics[width=1.0\textwidth]{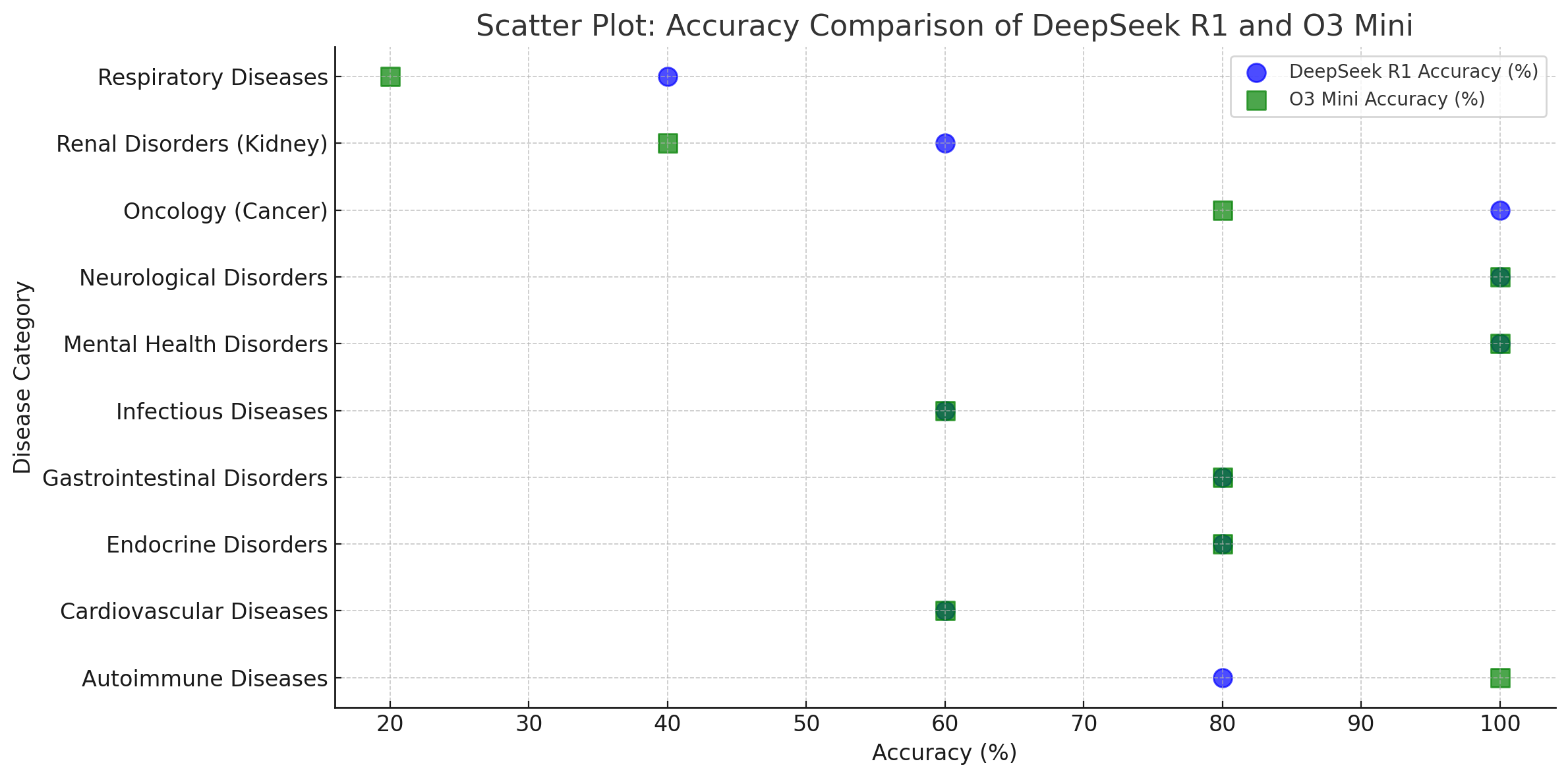}
    \caption{Scatter Plot: Accuracy Comparison of DeepSeek R1 and O3 Mini.}
    \label{fig:model_performance}
\end{figure}

The scatter plot in Figure~\ref{fig:model_performance} visually represents the classification accuracy of both models across different disease categories. The results highlight that DeepSeek R1 consistently outperformed O3 Mini in most disease classifications, particularly in \textbf{Mental Health, Neurological Disorders, and Oncology (Cancer)}, where it reached 100\% accuracy. This suggests that DeepSeek R1 is highly proficient in diagnosing conditions that have well-defined symptomatology and established medical literature. 

\begin{table}[h]
    \centering
    \renewcommand{\arraystretch}{1.3} 
    \begin{tabular}{|l|c|c|}
        \hline
        \rowcolor[gray]{0.85} 
        \textbf{Metric} & \textbf{DeepSeek R1 (\%)} & \textbf{O3 Mini (\%)} \\ 
        \hline
        \rowcolor[gray]{0.92} 
        Disease-Level Accuracy & \textbf{76.00}\% & \textbf{72.00}\% \\ 
        Category-Level Accuracy & 88.00\% & 78.00\% \\ 
        \rowcolor[gray]{0.92} 
        Overall Accuracy & 82.00\% & 75.00\% \\ 
        \hline
    \end{tabular}
    \caption{Comparison of Accuracy Metrics for DeepSeek R1 and O3 Mini.}
    \label{tab:model_comparison}
\end{table}

\subsection{Category-Specific Accuracy Analysis}

A deeper examination of performance across disease categories (Table~\ref{tab:model_comparison}) reveals distinct strengths and weaknesses. DeepSeek R1 demonstrated superior accuracy in diagnosing \textbf{Mental Health, Neurological Disorders, and Oncology (Cancer)}, reaching 100\% accuracy in these domains. This suggests that the model is highly effective at recognizing symptom patterns for well-documented diseases with clear diagnostic criteria. 

O3 Mini, on the other hand, exhibited a unique advantage in \textbf{Autoimmune Diseases}, achieving 100\% accuracy compared to DeepSeek R1’s 80\%. Autoimmune diseases often have diverse and overlapping symptom presentations, making accurate classification challenging. The performance of O3 Mini in this category suggests that it may be particularly effective at recognizing subtle variations in symptoms that characterize these conditions.

Despite these successes, both models exhibited challenges in classifying \textbf{Respiratory Diseases}, where DeepSeek R1 achieved 40\% accuracy and O3 Mini trailed at 20\%. This lower performance suggests that both models may struggle with diseases that have highly overlapping symptoms, such as differentiating between asthma and chronic obstructive pulmonary disease (COPD). Additionally, in \textbf{Cardiovascular, Infectious, and Renal Disorders}, both models performed moderately, indicating areas where further model fine-tuning or the incorporation of additional medical knowledge could improve performance.

\subsection{Confidence Score Analysis}

In addition to accuracy, the confidence levels of model predictions were analyzed. The confidence score distribution, as detailed in Table~\ref{tab:confidence_distribution_percentage}, revealed that DeepSeek R1 provided high-confidence predictions in 92\% of cases, whereas O3 Mini exhibited high confidence in 68\% of cases. Predictions classified as medium confidence accounted for 8\% in DeepSeek R1 and 32\% in O3 Mini. Notably, neither model generated low-confidence predictions, which suggests that both LLMs have a high degree of internal certainty in their classifications. 

\begin{figure}[h]
    \centering
    \includegraphics[width=1.0\textwidth]{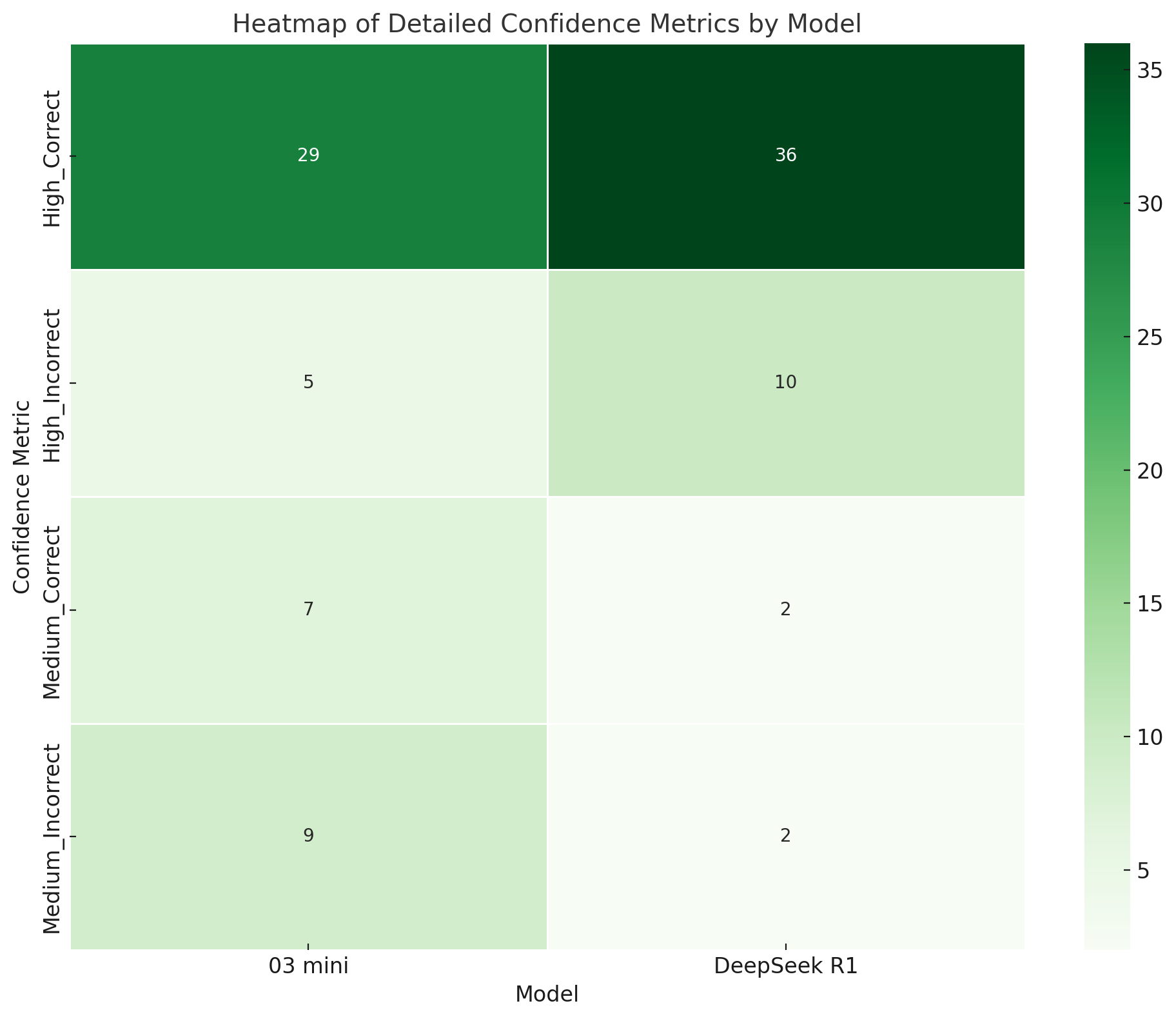}
    \caption{Heatmap of Confidence Score Metrics for DeepSeek R1 and O3 Mini.}
    \label{fig:confidence_analysis}
\end{figure}

The heatmap in Figure~\ref{fig:confidence_analysis} further highlights how confidence correlates with correctness. DeepSeek R1’s higher proportion of high-confidence correct classifications suggests a more reliable decision-making process, reducing the likelihood of uncertainty in clinical applications. Conversely, O3 Mini exhibited a greater proportion of medium-confidence predictions, indicating that while it provides reasonable accuracy, it often does so with less certainty.

\begin{table}[h]
    \centering
    \renewcommand{\arraystretch}{1.3} 
    \begin{tabular}{|l|c|c|}
        \hline
        \rowcolor[HTML]{D3D3D3} 
        \textbf{Confidence Level} & \textbf{DeepSeek R1 (\%)} & \textbf{O3 Mini (\%)} \\ 
        \hline
        \rowcolor[HTML]{F5F5F5} 
        High & \textbf{92.00}\% & \textbf{68.00}\% \\ 
        Medium & 8.00\% & 32.00\% \\ 
        \rowcolor[HTML]{F5F5F5} 
        Low & 0.00\% & 0.00\% \\ 
        \hline
    \end{tabular}
    \caption{Confidence Score Distribution (Percentage) for DeepSeek R1 and O3 Mini.}
    \label{tab:confidence_distribution_percentage}
\end{table}

\subsection{Comparative Insights}

From the overall evaluation, it is evident that DeepSeek R1 consistently outperforms O3 Mini in terms of classification accuracy and confidence. The higher confidence levels and improved classification accuracy suggest that DeepSeek R1 may be a more suitable candidate for real-world medical applications requiring high certainty in diagnoses. However, O3 Mini's strong performance in Autoimmune Diseases suggests that different models may specialize in specific medical domains.

A key area requiring further improvement is Respiratory Disease classification, where both models struggled. Future work could involve fine-tuning these models with specialized datasets or hybrid approaches combining different LLM architectures for better disease-specific performance.

These findings provide critical insights into the strengths and limitations of LLM-based medical diagnostics, offering a roadmap for future advancements in AI-driven healthcare.

\section{Discussion}

Our evaluation of DeepSeek R1 and O3 Mini demonstrates that LLMs can provide promising diagnostic support in disease classification, offering a new dimension to the way clinical data is processed and interpreted. The ability of these models to analyze complex symptom patterns and generate diagnostic predictions suggests that they have the potential to augment traditional diagnostic methods, thereby enhancing clinical decision-making. However, the study also reveals several important considerations and challenges that must be addressed before such systems can be widely adopted in clinical practice. Critical issues such as model interpretability, integration with existing healthcare systems, and the need for robust validation across diverse patient populations highlight the careful and methodical approach required for transitioning these promising tools from research to real-world application.

\subsection{Ethical Concerns}

The deployment of LLMs in medical diagnostics raises several ethical issues that must be addressed to ensure responsible usage. Our results indicate that DeepSeek R1 consistently outperforms O3 Mini in most disease categories—with overall accuracy of 82\% compared to 75\%—yet both models exhibit challenges in specific areas, such as Respiratory Disease classification (40\% and 20\%, respectively). These discrepancies underscore the risk that biases in training data may lead to uneven diagnostic performance across different conditions and, potentially, across diverse patient populations\cite{info15020092}\cite{he2023}.

Fairness is a critical concern. If certain demographics or disease presentations are underrepresented, the LLMs might perform well overall but fail for specific groups or conditions. Transparency in model development, rigorous validation on diverse datasets, and continuous monitoring in clinical settings are essential to address such disparities\cite{mcduff2023}.

Furthermore, the “black-box” nature of these models, as seen in our confidence score analysis—where DeepSeek R1 exhibited high confidence in 92\% of cases versus 68\% for O3 Mini—highlights the need for explainability. Clinicians must understand the rationale behind model predictions, particularly when incorrect diagnoses occur. Establishing clear guidelines that position LLMs as decision-support tools rather than replacements for clinical judgment will be crucial to ensure accountability.

Patient privacy and data security are paramount in any diagnostic process, especially when leveraging LLM-based platforms for medical decision-making. Strict adherence to data protection regulations, such as HIPAA and GDPR, is essential to safeguard sensitive patient information. Ensuring compliance with these standards not only protects against unauthorized access and misuse but also upholds the trust and confidentiality that are foundational to clinical practice\cite{nature2024}\cite{ai6010013}.

\subsection{Limitations}

While the quantitative metrics are encouraging, several limitations must be acknowledged. The dataset used in this study, although extensive, may not capture the full complexity of real-world clinical scenarios. For instance, our analysis revealed high accuracy in domains like Mental Health, Neurological Disorders, and Oncology, yet both models struggled with Respiratory Diseases—a reflection of potential gaps in the dataset or the inherent difficulty of classifying conditions with overlapping symptoms.

Additionally, the exclusive reliance on textual symptom descriptions limits diagnostic precision. Clinical diagnoses typically require multi-modal inputs such as imaging, laboratory tests, and patient histories. Our current approach, which does not integrate these modalities, might lead to oversimplified assessments that do not fully capture the nuances of patient presentations.

Moreover, while our evaluation framework includes quantitative metrics such as disease-level and category-level accuracy, these measures do not entirely capture the clinical interpretability of model outputs. The confidence scores provide a basic indication of reliability, yet the underlying decision-making process remains opaque. This limitation could hinder clinical trust and acceptance.

\subsection{Future Directions}

The findings of this study suggest several avenues for future research. Enhancing model generalizability by incorporating larger, more diverse datasets is imperative. Future work should aim to update training data continuously to reflect evolving clinical practices and to cover a broader spectrum of patient demographics and disease presentations.

The integration of multi-modal data is another promising direction. Combining textual symptom analysis with other sources—such as medical imaging, laboratory results, and comprehensive patient histories—could yield a more holistic diagnostic tool, potentially addressing the observed deficiencies in areas like Respiratory Disease classification.

Improving model interpretability remains a critical focus. Future research should explore explainable AI techniques that clarify the decision-making process behind LLM predictions. Techniques such as attention visualization or rule extraction could help elucidate why models like DeepSeek R1 and O3 Mini assign certain confidence levels, thereby building clinician trust.

Lastly, rigorous clinical validation through prospective studies and controlled trials is essential. Real-world evaluations will help confirm the robustness and reliability of these LLM-based systems and determine how best to integrate them into clinical workflows, ensuring that they complement and enhance, rather than replace, clinical expertise. Collectively, addressing these ethical, technical, and practical challenges will be vital for harnessing the full potential of LLMs in medical diagnostics and ultimately improving patient outcomes.

\section{Conclusion}

This study evaluated the diagnostic performance of DeepSeek R1 and O3 Mini in disease classification across multiple categories using a rigorously structured dataset and a comprehensive quantitative evaluation framework. Our results demonstrate that LLMs can provide promising diagnostic support in medical applications. Specifically, DeepSeek R1 outperformed O3 Mini in terms of overall accuracy and confidence levels, indicating its superior ability to analyze complex symptom patterns. However, O3 Mini showed a notable strength in Autoimmune Disease classification, suggesting that different models may have specialized advantages in certain clinical domains.

The study also identified critical areas for improvement, particularly in the classification of Respiratory Diseases, where both models underperformed. These findings highlight the need for continued refinement and adaptation of LLM-based diagnostic tools to address the inherent challenges of overlapping symptomatology and data variability. Additionally, the importance of ethical considerations—including transparency, fairness, and compliance with data protection regulations—was underscored as essential for the responsible integration of LLMs into clinical practice.

In conclusion, our research contributes valuable insights to the growing body of work on LLM-based medical diagnostics. As these models continue to evolve, ensuring rigorous clinical validation and the development of more interpretable, equitable diagnostic tools will be paramount. This will not only enhance diagnostic accuracy but also build greater trust among clinicians and patients, ultimately advancing global healthcare outcomes.

\bibliography{references} 

\bibliographystyle{unsrt}  

\nocite{*}

\section*{Abbreviations}
The following abbreviations are used in this manuscript:
\begin{itemize}
    \item AI: Artificial Intelligence
    \item LLM(s): Large Language Model(s)
    \item HIPAA: Health Insurance Portability and Accountability Act
    \item GDPR: General Data Protection Regulation
    \item EHR: Electronic Health Record
    \item NLP: Natural Language Processing
\end{itemize}

\clearpage 
\appendix
\section{Accuracy Comparison of LLMs Across Disease Categories}

\begin{table}[h]
    \centering
    \begin{tabular}{|l|c|c|}
        \hline
        \textbf{Disease Category} & \textbf{DeepSeek R1 Accuracy (\%)} & \textbf{O3 Mini Accuracy (\%)} \\ 
        \hline
        Autoimmune Diseases & 80.00 & 100.00 \\
        Cardiovascular Diseases & 60.00 & 60.00 \\
        Endocrine Disorders & 80.00 & 80.00 \\
        Gastrointestinal Disorders & 80.00 & 80.00 \\
        Infectious Diseases & 60.00 & 60.00 \\
        Mental Health Disorders & 100.00 & 100.00 \\
        Neurological Disorders & 100.00 & 100.00 \\
        Oncology (Cancer) & 100.00 & 80.00 \\
        Renal Disorders (Kidney) & 60.00 & 40.00 \\
        Respiratory Diseases & 40.00 & 20.00 \\
        \hline
    \end{tabular}
    \caption{Accuracy Comparison of DeepSeek R1 and O3 Mini Across Disease Categories.}
    \label{tab:disease_category_accuracy_updated}
\end{table}

\end{document}